\documentclass[sigconf]{acmart}
\AtBeginDocument{%
  \providecommand\BibTeX{{%
    \normalfont B\kern-0.5em{\scshape i\kern-0.25em b}\kern-0.8em\TeX}}}

\setcopyright{acmcopyright}
\copyrightyear{2023}
\acmYear{2023}
\acmDOI{XXXXXXX.XXXXXXX}

\acmConference[CIKM '23]{Proceedings of the 32nd ACM International Conference on Information and Knowledge Management}{October 21--25, 2023}{Birmingham, United Kingdom}
\acmBooktitle {Proceedings of the 32nd ACM International Conference on Information and Knowledge Management (CIKM '23), October 21--25, 2023, Birmingham, United Kingdom}
\acmPrice{15.00}
\acmISBN{978-1-4503-XXXX-X/18/06}

\acmSubmissionID{963}

\copyrightyear{2023} 
\acmYear{2023} 
\setcopyright{acmlicensed}\acmConference[CIKM '23]{Proceedings of the 32nd ACM International Conference on Information and Knowledge Management}{October 21--25, 2023}{Birmingham, United Kingdom}
\acmBooktitle{Proceedings of the 32nd ACM International Conference on Information and Knowledge Management (CIKM '23), October 21--25, 2023, Birmingham, United Kingdom}
\acmPrice{15.00}
\acmDOI{10.1145/3583780.3614671}
\acmISBN{979-8-4007-0124-5/23/10}

\begin{document}

\title[A Spatial-Temporal Attention Network for Logistics Delivery Timely Rate Prediction in Anomaly Conditions]{DeepSTA: A Spatial-Temporal Attention Network for Logistics Delivery Timely Rate Prediction in Anomaly Conditions}

\author{Jinhui Yi}
\email{yijh21@mails.tsinghua.edu.cn}
\orcid{0000-0002-2339-2698}
\affiliation{%
  \institution{BNRist, Department of Electronic Engineering, Tsinghua University}
  \institution{JD Logistics}
  \streetaddress{30 Shuangqing Rd}
  \state{Beijing}
  \country{China}
  \postcode{100080}
}

\author{Huan Yan$^{\ast}$}
\email{yanhuan@tsinghua.edu.cn}
\affiliation{%
  \institution{BNRist, Department of Electronic Engineering, Tsinghua University}
  \streetaddress{30 Shuangqing Rd}
  \city{Beijing}
  \country{China}
}

\author{Haotian Wang}
\email{wanghaotian18@jd.com}
\affiliation{%
  \institution{JD Logistics}
  \city{Beijing}
  \country{China}
}

\author{Jian Yuan}
\email{jyuan@tsinghua.edu.cn}
\affiliation{%
 \institution{BNRist, Department of Electronic Engineering, Tsinghua University}
 \streetaddress{30 Shuangqing Rd}
 \state{Beijing}
 \country{China}
 }

 \author{Yong Li}
 \email{liyong07@tsinghua.edu.cn}
\affiliation{%
 \institution{BNRist, Department of Electronic Engineering, Tsinghua University}
 \streetaddress{30 Shuangqing Rd}
 \state{Beijing}
 \country{China}
 }

\thanks{$*$ Corresponding author. Email: yanhuan@tsinghua.edu.cn}
\renewcommand{\shortauthors}{Jinhui Yi, Huan Yan, Haotian Wang, Jian Yuan, \& Yong Li.}

\begin{abstract}
Prediction of couriers' delivery timely rates in advance is essential to the logistics industry, enabling companies to take preemptive measures to ensure the normal operation of delivery services. This becomes even more critical during anomaly conditions like the epidemic outbreak, during which couriers' delivery timely rate will decline markedly and fluctuates significantly.
Existing studies pay less attention to the logistics scenario.
Moreover, many works focusing on prediction tasks in anomaly scenarios fail to explicitly model abnormal events, \textit{e.g.}, treating external factors equally with other features, resulting in great information loss. Further, since some anomalous events occur infrequently, 
traditional data-driven methods perform poorly in these scenarios.
To deal with them, we propose a deep spatial-temporal attention model, named DeepSTA.
To be specific, to avoid information loss, we design an anomaly spatio-temporal learning module that employs a recurrent neural network to model incident information. Additionally, we utilize Node2vec to model correlations between road districts, and adopt graph neural networks and long short-term memory to capture the spatial-temporal dependencies of couriers. 
To tackle the issue of insufficient training data in abnormal circumstances, we propose an anomaly pattern attention module that adopts a memory network for couriers' anomaly feature patterns storage via attention mechanisms. 
The experiments on real-world logistics datasets during the COVID-19 outbreak in 2022 show the model outperforms the best baselines by 12.11\% in MAE and 13.71\% in MSE, demonstrating its superior performance over multiple competitive baselines.
\end{abstract}
%






\maketitle

\vspace{-4mm}
\section{INTRODUCTION}
\label{Sec:introduction}
The logistics companies such as JD Logistics and SF Express, provide hundreds of millions of users with logistics services including package delivery and pick-up. 
The delivery timely rate of couriers, which represents the percentage of parcels delivered on time by each courier, serves as a measure of work efficiency and has a direct impact on customers' experience. This metric is of significant importance to the industry, as it reflects the quality of service provided by couriers and ultimately influences customer satisfaction.

\begin{figure}[htbp]
\centering
\includegraphics[width=\linewidth]{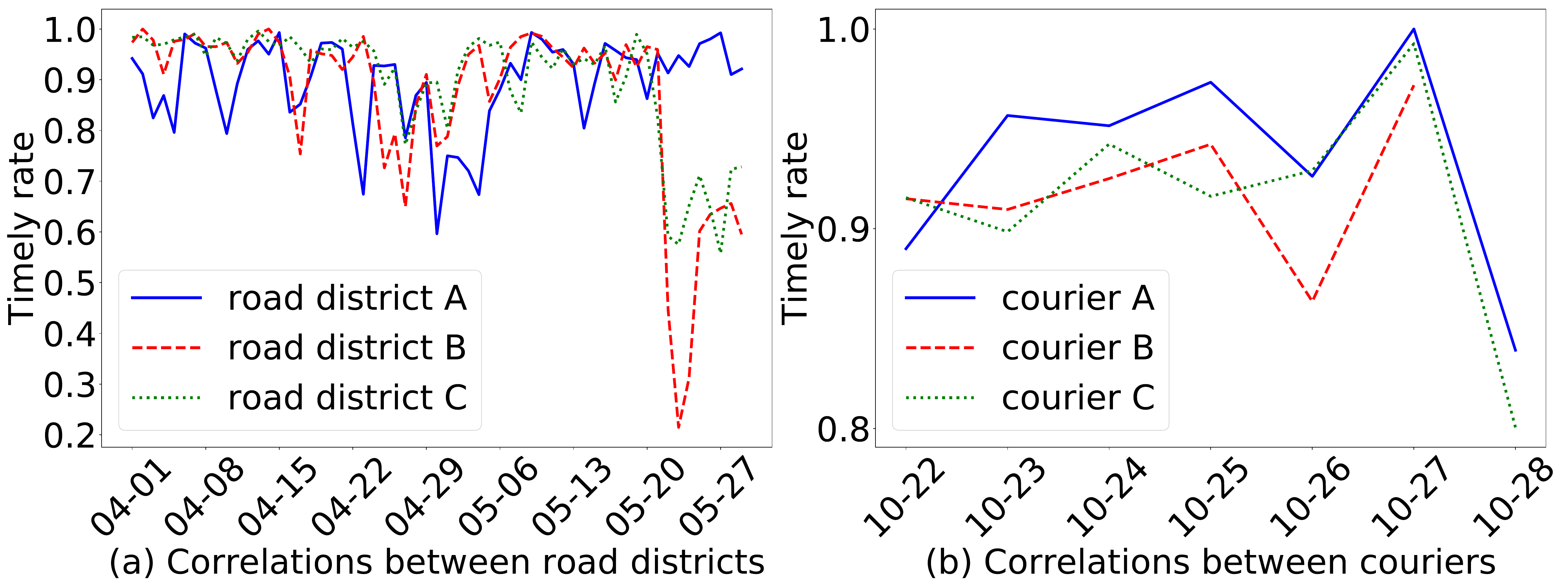}
\vspace{-5mm}
\caption{(a) District A is far from B and C which are adjacent. In May 2022, the timely rate on B and C dropped synchronously due to the regional epidemic lockdown.
(b) On October 28th, courier B was absent due to the lockdown, and his parcels were reassigned to A and C.
} 
\label{relation}
\vspace{-6mm}
\end{figure}

Predicting the delivery timely rate in advance is crucial to logistics companies for resource allocation and work assignment~\cite{ren2022deepexpress}, which is even more necessary in anomaly circumstances like epidemic outbreaks and lockdowns during COVID-19~\cite{xia2023predict,fang2019mac}.
As shown in Figure~\ref{relation}, the delivery timely rate is a real number between 0 and 1, and companies typically require this metric to be no lower than 0.9. 
However, during the small-scale outbreak of COVID-19 in May and October of 2022, in Beijing, due to factors such as regional lockdowns and staff infections which hinder the normal work of couriers, the delivery timely rate dropped significantly below 0.8, resulting in a substantial increase in customer complaints. 
If the companies can early predict  which couriers tend to have lower timely rates, then they can take interventions such as allocating crowdsourcing workers to help those couriers, thereby ensuring service quality and gaining a competitive advantage.
\vspace{-3mm}

However, existing research on online businesses has paid limited focus on the express scenario, with more attention to the food takeaway scenario~\cite{abdelraouf2022sequence,gao2021deep,zhu2020order}.
These works usually estimate vehicles' arrival time in normal circumstances using fine-grained data like real-time position and historical routes.  
However, in abnormal situations, such information is difficult to collect in advance due to factors like regional lockdowns, which makes it difficult to apply those route-based real-time arrival estimation approaches.
Considering these obstacles, the industry typically adopts conventional time series forecasting approaches ~\cite{contreras2003arima,bai2018empirical}. 
These data-driven methods treat both anomalous and normal information equally, resulting in the loss of anomalous information. Moreover, these approaches perform poorly during anomalous events because of insufficient training data. Thus, we need to address two key challenges: 

a) \textbf{Complex impact of anomaly events on delivery.} 
To simulate the impact of anomaly events, a common way is to use external factors like newly confirmed cases as the model input~\cite{hu2022black,li2023traffic}. 
However, most of the existing works treat the external information and other information equally, by directly concatenating them together, which causes the loss of external information.
In addition, previous works on anomaly detection ~\cite{chen2022graphad,xie2020deep} show that mining correlations between individuals through graph algorithms can contribute to the model performance. However, it is challenging for conventional approaches to model such correlations directly. 
As shown in Figure~\ref{relation}, there are two types of correlations in the express delivery, \emph{i.e.}, the correlations between road districts, and the correlations between couriers.
The distance-based correlations between road districts which are derived from the city, prove to be more obvious due to the geographically concentrated distribution of confirmed cases and neighborhood lockdowns. 
For the latter, if couriers encounter unexpected factors such as personal illness or lockdowns, their packages will be reassigned to other couriers with respect to the team arrangement mode, whose timely rate may be impacted due to the increased workload.
Thus, it is difficult to capture the complex impact of anomaly incidents.

b) \textbf{Lack of sufficient training data.} 
Under normal circumstances, the delivery timely rate is usually above 0.9. However, it will drop significantly and fluctuate dramatically during anomalous events such as epidemic outbreaks, with samples lower than 0.8 or even 0.6. 
Nevertheless, before the announcement of the lifting of COVID-19 measures in late November 2022, the spread of COVID-19 was mostly under control, resulting in the lack of anomalous data.
Due to the diversity gap between the training dataset of normal situations and the testing dataset of abnormal situations, the traditional data-driven approaches have limitations in effectively capturing the complex patterns and dependencies of the abnormal features, leading to overfitting to the training dataset and poor performance facing anomalous events.
Especially, anomaly events such as COVID-19 outbreaks after the loosening of control policies which occur with low frequency or high volatility, will make these methods' performance even worse.

To address the above challenges, 
we propose a spatial-temporal attention neural network named DeepSTA, consisting of the anomaly spatio-temporal learning module and the anomaly pattern attention module.
The anomaly spatio-temporal learning module is designed to tackle the first challenge. Firstly, it utilizes Node2vec to represent correlations between road districts. Secondly, Graph Convolutional Neural Network (GCN) is employed to extract correlations between couriers. Thirdly, Long Short-Term Memory (LSTM) is utilized to capture temporal dependencies. Lastly, a Recurrent Neural Network (RNN) is employed to model the impact of external factors and extract anomaly information, thus avoiding information loss.
To tackle the second challenge, the anomaly pattern attention module employs a memory network to capture the patterns and dependencies of features in abnormal situations via the attention mechanism, thereby compensating for the lack of training data.

The main contributions of this work are summarized below:

1) 
To the best of our knowledge, we are the first to propose a deep spatio-temporal attention model on timely rate prediction in anomaly conditions.

2) 
To overcome the difficulty of the lack of training data, we design a memory network to learn patterns of data.
The memory network learns the patterns of features 
in anomaly situations via external memory.

3) 
To better model the anomaly information, we employ an RNN network for external information processing, which models the impact of external factors separately.

4) 
Experiments on a real-world logistics dataset suggest that our approach outperforms the best baselines by 12.11\% in MAE and 13.71\% in MSE. The model has been deployed online in JD Logistics.


\vspace{-3mm}
\section{RELATED WORKS}
\label{Sec:related_work}
\subsection{Online Business Prediction Problem}
\label{sec:Arrival}
Express and food takeaway are the two main applications of online commerce, which involve a large amount of spatial-temporal information and can be utilized to enhance the performance of the model.
The literature on the express service is relatively scarce, with most of them focusing on forecasting the logistics situation in larger regions (province or country) in the long (monthly or annual level) term using macroeconomic data (such as GDP, import and export data)~\cite{huang2021regional,yu2020research,yan2019research,yin2016data,liu2022pred}. These studies neglect to explore the correlations between individuals in the express service, and are hard to utilize for couriers' level timely rate forecasting. 

In comparison, existing studies on online businesses have paid more attention to the food takeaway scenario, which usually extract information from the couriers’ time variant information like route and location~\cite{chen2022hseta,gao2021deep,ruan2020doing,zhu2020order,guo2022wepos,zhang2015feeder,yang2018sharededge} 
for arrival time estimation, by using the graph to represent spatial relations ~\cite{chen2022graphad,fang2021traffic,yu2020forecasting} and utilizing the RNN to learn temporal dependencies ~\cite{gao2021deep,wang2019pedestrian,wen2021package,wu2018graph,yao2018modeling,zhao2019t}. 
However, 
such approaches are hard to apply directly in anomaly circumstances as stated in section ~\ref{Sec:introduction}. 
Still, inspired by those work, we utilize Node2vec~\cite{grover2016node2vec} and GCN~\cite{kipf2016semi} to model spatial correlations and employ LSTM~\cite{hochreiter1997long} to capture temporal dependencies.
\vspace{-3mm}
\subsection{Time Series Prediction Methods}
The timely rate prediction is a typical time series prediction task.
Traditional time series prediction methods including 
moving average (MA)~\cite{box1970distribution}, 
and auto-regressive integrated moving average (ARIMA)~\cite{contreras2003arima},
typically learn the patterns of changes in time series exclusively from the time series itself and do not consider the interactions between the time series and other relevant information.
Deep learning-based methods including the LSTM, 
the Sequence to Sequence (Seq2seq)~\cite{sutskever2014sequence} and  
the Temporal Convolutional Network (TCN)~\cite{bai2018empirical}, 
mainly focus on capturing the temporal dependencies of sequence, omitting the spatial correlations which will enhance the prediction performance.
\vspace{-3mm}
\subsection{Anomaly Learning}
\label{sec:Memory Networks}
Unlike periodic events like holidays, anomaly events
occur very rarely, resulting in a scarcity of corresponding training data. This makes it difficult to apply historical data-based methods such as transfer learning~\cite{hu2022black} or sequences similarity~\cite{zhu2021hybrid,zhang2019decomposition}. 
However, recent works~\cite{wang2022event,li2023traffic} have leveraged the memory network to learn the characteristics of anomalous events.
The memory network is first introduced in natural language processing for Question Answering~\cite{chaudhari2021attentive}
via an external memory for information retrieval.
Recently, it has been adopted in time series forecasting~\cite{yao2019learning,tang2020joint,li2023traffic}. 
In this work, the memory network is utilized to address the challenge of insufficient training data in anomaly environments.

\vspace{-3mm}
\section{PROBLEM FORMULATION}
The logistics company divides the city into multiple road districts based on the real-road network. 
For the delivery timely rate prediction in anomaly conditions, suppose there are $N$ couriers 
responsible for $M$ road districts in all. $i(i=1,…, N)$ and $j(j=1,…, M)$ denote the indices of couriers and road districts, respectively. For courier $i$, his/her known information a day before timestamp $t$ is denoted by $X_i^t$, which includes logistics information, weather forecasting information, date information and anomaly external factors. And his/her delivery timely rate on timestamp $t$ is denoted by $Y_i^t$ which is a real number between (0,1), 
where $X_i^t \in R^{D}$ and $Y_i^t \in R$, with $D$ being the dimension of the information vector of each day. 
Given the information of $T$ days ahead of $t$,
our goal is to predict the $N$ couriers' delivery timely rates of the next day $t$, \emph{i.e.}, 
\vspace{-1.6mm}
\begin{equation}
\Phi: \{X_i^t,X_i^{t-1},\cdots,X_i^{t-T} \} \rightarrow Y_i^t, 
\end{equation}
\vspace{-0.8mm}
where $\Phi$ is the mapping regression function to learn. 
In particular, we focus on predicting timely rates in anomaly condition, \emph{i.e.}, 
the training set consists of data from normal situations and the testing set consists of data from anomaly situations.
\vspace{-2mm}
\section{METHOD}\label{Sec:method}
\subsection{The Proposed Framework}
\begin{figure*}[htbp]
\includegraphics[width=\linewidth]{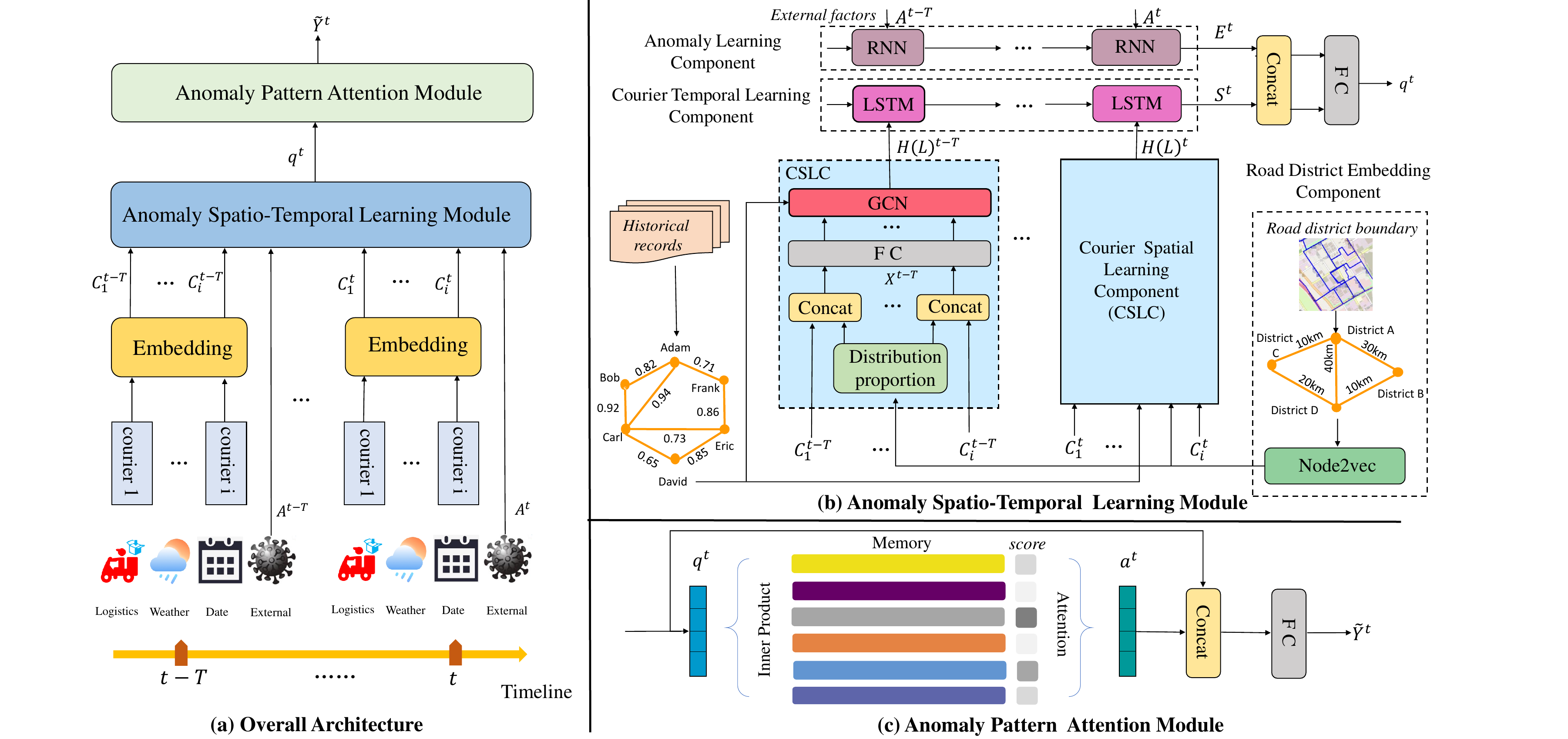}
\caption{ Overall architecture of DeepSTA.} 
\vspace{-3mm}
\label{figmodel}
\end{figure*}

An overview of DeepSTA is shown in Figure~\ref{figmodel}, which consists of two modules: the anomaly spatio-temporal learning module and the anomaly pattern attention module. 
The anomaly spatio-temporal learning module is comprised of four components.
First, the road district embedding component models the spatial dependencies of road districts, by constructing a graph of road districts  and utilizing Node2vec~\cite{grover2016node2vec} to generate the embedding of road districts.
Second, the courier spatial learning component captures correlations between couriers, by constructing the graph of couriers with historical logistics data and fusing all features on each timestamp 
via GCN.
Third, the courier temporal learning component learns the temporal dependencies of the logistics sequences. For each courier, a sequence 
of length $T+1$, which includes both the historical information of $T$ previous timestamps and the information of day $t$ to be predicted, will be transmitted into LSTM.
Fourth, the anomaly learning component encodes external factors via RNN. 
The anomaly pattern attention module concatenates the output of the anomaly learning component and the courier temporal learning component together, which will be fed into a memory network for information retrieval to generate final prediction results.

\newlength{\mywidth}
\newlength{\myheight}
\settowidth{\mywidth}{\includegraphics{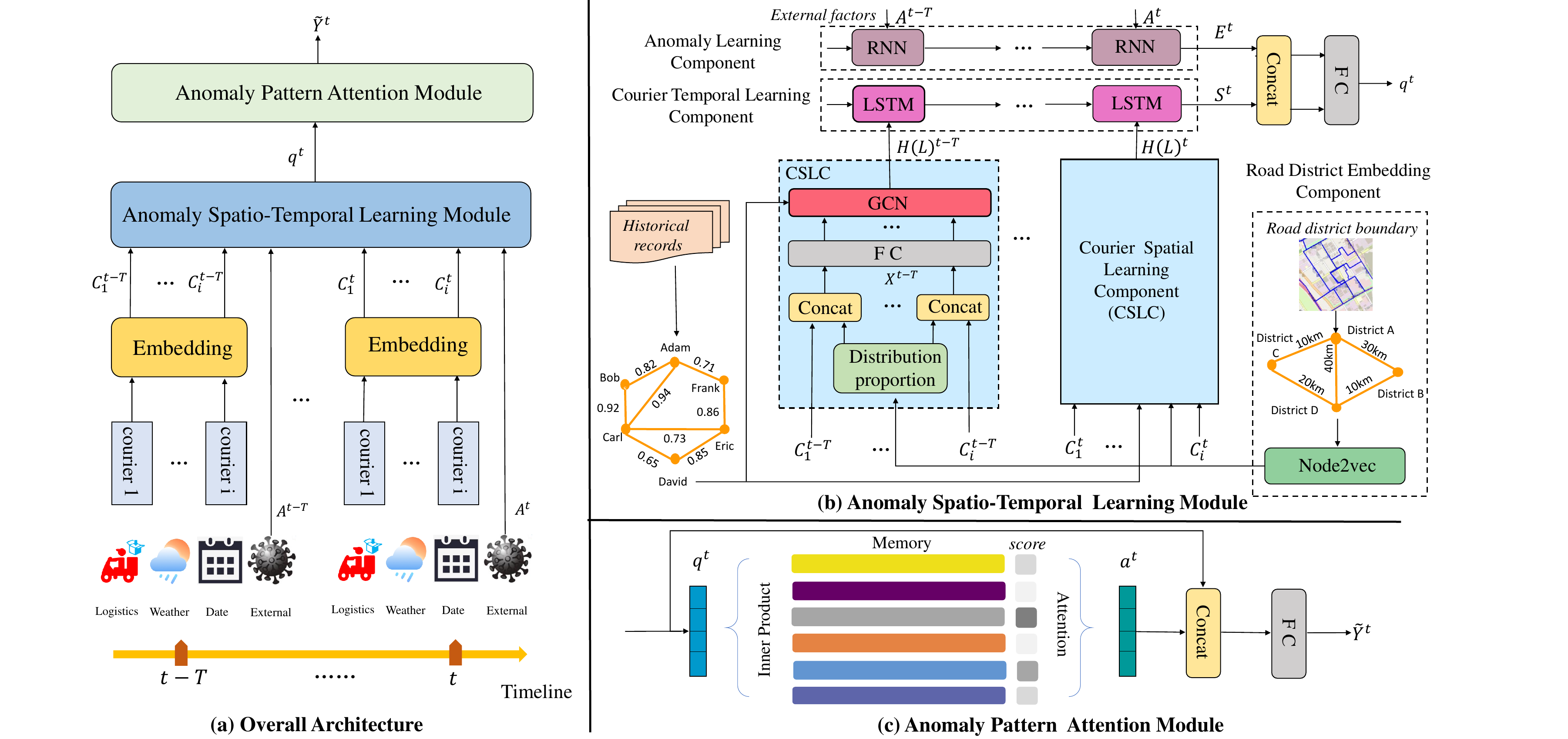}}
\settoheight{\myheight}{\includegraphics{figures/model_newest.pdf}}
\vspace{-1mm}
\vspace{-2mm}


\subsection{Anomaly Spatio-Temporal Learning Module}
This module is designed to capture spatio-temporal dependencies in anomaly scenarios, consisting of the following four components:
\vspace{-6mm}
\subsubsection{Road District Embedding Component}
This component is designed to capture the spatial patterns of road districts. Each road district is surrounded by a polygonal boundary. We first calculate the coordinates of each road district’s centroid point. Then we match these points to the nodes of the real-road network using OpenStreetMap~\cite{haklay2008openstreetmap}. Thus we construct a directed fully connected graph $G_r = ( V , E )$, where $V$ is the set of vertices corresponding to road districts respectively, and $E$ is the set of edges which is the shortest path connecting two nodes on the road network. The weight of each edge is set as the reciprocal of the shortest distance between neighboring vertices, with the element of the adjacency matrix set to the edge weight respectively.
As shown in Figure~\ref{relation}, in abnormal situations, 
due to factors such as lockdowns, the closer the distance between road districts, the stronger their correlations becomes.
To model such correlations, an intuitive approach is using graph algorithms, among which graph neural networks are usually adopted for supervised tasks with labeled data.
Since road districts do not have ground truth labels, like dish type of restaurants in food takeaway scenario ~\cite{chen2022graphad}. 
Therefore, we adopt the graph embedding method, Node2vec, which is widely used in unsupervised tasks, to learn spatial representations of nodes in $G_r$. The 128-dimensional output embedding ${Road_j}$, where $j$ denotes road district $j$, will be part of the model input in the next component.
\vspace{-2.5mm}
\subsubsection{Courier Spatial Learning Component}
This component aims to capture correlations between couriers, which consists of two parts: coarse-grained information estimation and couriers correlation representation. 
The first part intends to tackle the difficulty of obtaining fine-grained information in anomaly conditions, which is currently unknown when forecasting the timely rate of the following day as mentioned in Section ~\ref{Sec:introduction}. The second part models the spatial correlation among couriers. 

\textbf{(1) Coarse-grained information estimation.}
Although the exact locations of packages within each road district vary from day to day and are unknown in advance, packages in the same district are usually assigned to the same courier. 
So for courier $i$, we calculate the proportion of his orders distributed in each district, and add the embedding of each district generated from the road district embedding component according to this ratio, 
$\mathbf{Emb_i^t}=p_{1i}^{t-1}*\mathbf{Road_1}+\cdots+p_{Mi}^{t-1}*\mathbf{Road_M}$, where $p_{ji}^{t-1}$ represents the proportion of courier $i$'s orders in district $j$ on timestamp $t-1$.
We then concatenate $Emb_i^t$ with $C_i^t$ which includes logistics features, weather forecasting features, and date features that are known a day ahead of $t$.
Then we pass these features through a fully connected layer to get the feature vector $X_i^t$. Thus all couriers' feature matrix on $t$ is generated via each courier's feature vector, \emph{i.e.}, $X^t=\{X_1^t,\cdots,X_i^t\}$.

\textbf{(2) Couriers correlation representation.}
Figure~\ref{relation} shows that couriers exhibit collaborative work behavior in abnormal circumstances with respect to the team arrangement mode.
Inspired by ~\cite{chen2022graphad} which calculates time-series similarity to represent the relation between entities, we model the correlations between couriers as an undirected graph $G_c$ whose nodes represent couriers. For each courier, we calculate the Pearson's correlation coefficient of delivery timely rate series from March 1st to April 1st between him/her and the rest of the couriers. For each pair, if their Pearson's correlation coefficient is nonnegative, their weight is set as the coefficient, otherwise the two nodes are treated as not connected. 
We then employ a GCN network with $L$ layers based on $G_c$, which projects the feature matrix $X^t$ into the output $\mathbf{H}(L)^t \in R^{D_g}$ on timestamp $t$.
\vspace{-3mm}
\subsubsection{Courier Temporal Learning Component}
This component adopts an LSTM network to learn temporal dependencies to enhance the model's performance facing anomaly events. Note that $T$ denotes 
the length of the input historical time sequence. At each timestamp ${t}$, the courier spatial learning component produces the output $\mathbf{H}(L)^{t}$. Subsequently, $\mathbf{H}(L)^{(t-T)},…, \mathbf{H}(L)^t$ are transmitted as input to the LSTM network, which will encode the historical temporal features up to the current timestamp $t$ into the output vector $\mathbf{S}^t \in R^{N \times D_s}$ on timestamp $t$, where $D_s$ is the projected dimension of each courier. 

\subsubsection{Anomaly Learning Component}
\begin{figure}[hptb]
\includegraphics[width=0.8\linewidth]{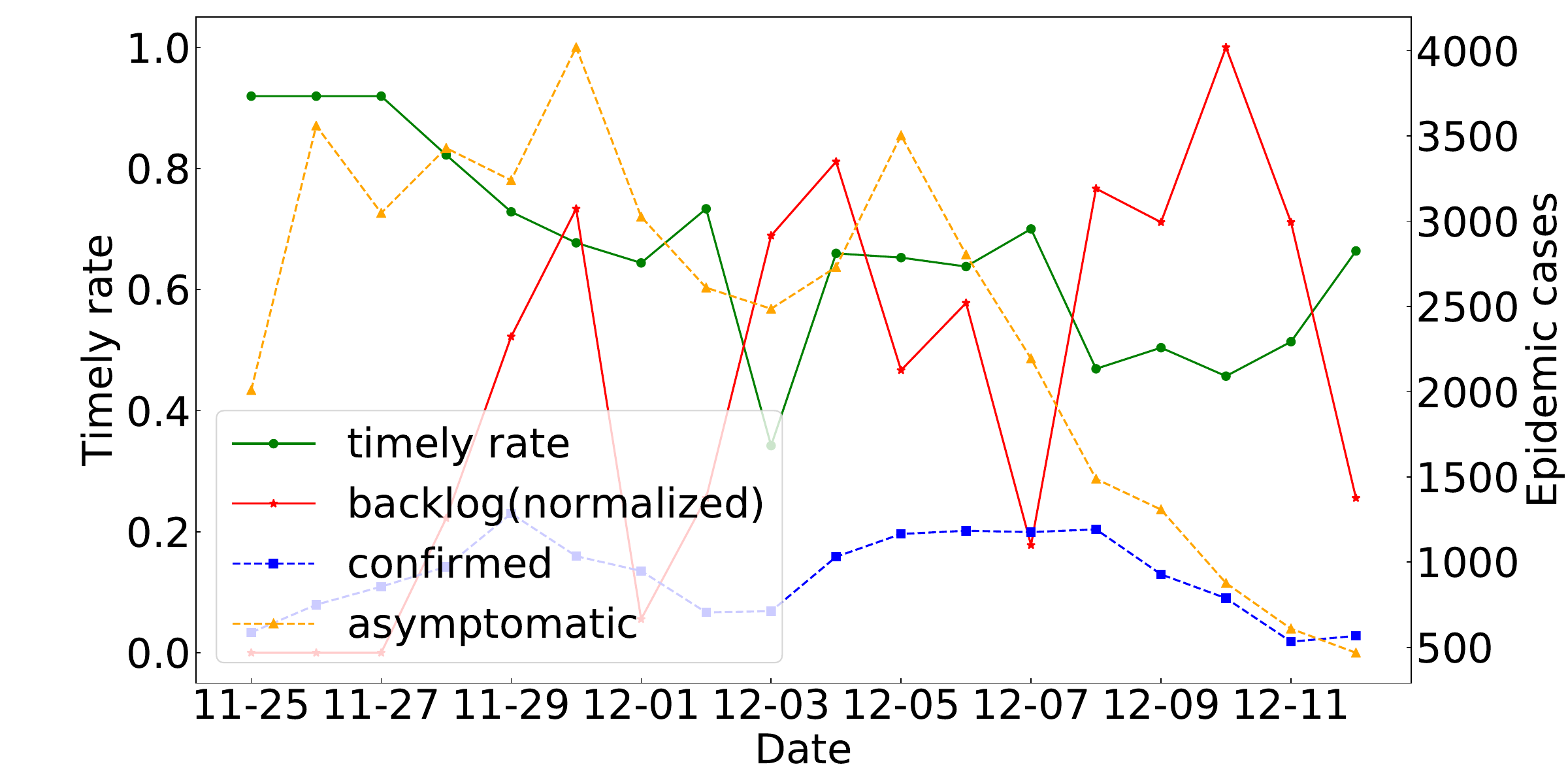}
\vspace{-4mm}
\caption{ Visualization of external factors.} \label{external}
\vspace{-4mm}
\end{figure}

This component is designed to handle anomaly information separately and learns the impact of external factors.
As anomaly events in 2022 were mainly caused by COVID-19,
we utilize the following 4-dimensional vector $\mathbf{A}^{t}$ as external factors: the daily number of newly confirmed cases in Beijing, the daily number of newly asymptomatic cases in Beijing, the daily number of newly confirmed cases in Tongzhou District of Beijing, and the daily backlog of undelivered packages for each courier. As shown in Figure~\ref{external}, the first three factors measure the changes in the external epidemic event. The fourth factor represents the number of undelivered packages for each courier on a given day. Based on our experience, the backlog of undelivered packages tends to increase continuously for several days after anomaly events occur. Moreover, 
the number of undelivered parcels on the previous day will affect  the timely rate of the following day.

\vspace{-1mm}
\vspace{-1mm}

Directly concatenating external factors with other features in the courier spatial learning component as the model input, will result in information loss. Therefore, we adopt a single-layer RNN to handle these factors separately whose input size is 4 and hidden size is 8. Thus, we transmit $\mathbf{A}^{t}$ into the RNN and get the 8-dimensional output vector $\mathbf{E}^{t}$.

\subsection{Anomaly Pattern Attention Module}
This module is designed to address the problem of the lack of training data via the attention mechanism in the memory network.
The external memory of the memory network learns and stores the patterns of anomaly events during the training stage.
We randomly initialize the external memory $\emph{M} \in R^{L_m \times D_m}$, where $L_m$ is the number of stored patterns which is set to $12$ and $D_m$ is the dimension of patterns which equals $64$. 
We first concatenate $\mathbf{S}^t$ with $\mathbf{E}^{t}$ to query the memory network: 
\begin{equation}
\mathbf{q}^{t}=\mathbf{W}_q[\mathbf{S}^t||\mathbf{E}^{t}]+\mathbf{b}_q,
\end{equation}
where $||$ denotes concatenation and $\mathbf{W}_q$ and $\mathbf{b}_q$ are trainable parameters. 
Then we calculate the similarity score between $\emph{M}$ and $\mathbf{q}^{t}$ via attention mechanism:
\begin{equation}
\mathbf{score}=Softmax(\emph{M} \cdot \mathbf{q}^{t}),
\end{equation}
where $\cdot$ denotes the inner product. Thus the representation vector is obtained by matrix multiplication:
\begin{equation}
\mathbf{a}^{t}=\mathbf{score} \times \emph{M},
\end{equation}
Then we concatenate $\mathbf{S}^t$, $\mathbf{E}^{t}$ and $\mathbf{a}^{t}$ together and feed them into a fully connected layer, whose dropout rate is 0.1 and activation function is sigmoid, to get the final prediction $\tilde{Y}^t$. 
\subsection{Model Training}
We train the whole model in an end-to-end manner using an MSE loss function:
\begin{equation}
MSE=\frac{1}{N} \sum_{i=1}^{N} {(Y_i^t- \tilde{Y}_i^t) }^2,
\end{equation}
where $Y^t$ is the true value of the delivery timely rate of $X^t$.



\section{EXPERIMENTS}\label{Sec:experiments}

\subsection{Experimental Setup}
\subsubsection{Dataset}
\vspace{-4mm}
\begin{table}[htbp]
\centering
\caption{Dataset Summary.}\label{Dataset Summary}
\vspace{-3mm}
\resizebox{1\linewidth}{!}{
\begin{tabular}{|c|c|c|c|}
\hline
\textbf{} &  \textbf{Training} & \textbf{Validation} & \textbf{Testing}\\
\hline
\textbf{\# of samples} &  110976 & 13600 & 14144\\
\textbf{\# of couriers} &  544 & 544 & 544\\
\textbf{\# of road districts} &  711 & 711 & 711\\
\textbf{\# of days} &  2, Apr. - 22, Oct. & 23, Oct. - 16, Nov. & 17, Nov. - 12, Dec.\\
\hline
\end{tabular}
}
\end{table}
\vspace{-2mm}
The dataset, as presented in Table~\ref{Dataset Summary}, is provided by JD Logistics, including the daily records of 544 couriers in Tongzhou District, Beijing, from March 1st, 2022, to December 12th, 2022. As each courier has a record for 286 days, the total number of samples is 155584.
Data from March 1st to April 1st are used to calculate the Pearson's correlation coefficient between couriers in the courier spatial learning component. 
Since before the Chinese government announced the loosening of COVID-19 control measures in late November 2022, the epidemic in China did not break out on a large scale. Therefore, the remaining data is divided into three parts in order, the first 80\% is used as the training set, and the respective 10\% is used as the validation and testing sets.

In detail, our datasets consist of five types: logistics information, geographical information, weather forecast information, date information and epidemic information. 
The logistics information records courier profiles(employee number, length of employment, courier age), the number and geographic distribution of orders from multiple services (delivery, merchant pick up and customer pick up), and corresponding daily timely rate of each service. 
We also calculate the daily backlog amount of delivery. 
The geographical information contains the boundary of road districts and the road network of the corresponding urban area from OpenStreetMap with the parameter "network type" set to "drive", indicating that the obtained roads are passable by vehicles.
The regional weather forecast information 
is acquired from ~\footnote{http://lishi.tianqi.com/}, including daily weather type forecasts, predicted highest, lowest and average temperatures. 
The date type information includes 
the day of the week and whether the day is a holiday.
The epidemic information is obtained from ~\footnote{https://news.sina.cn/zt\_d/yiqing0121}.
Among them, numerical features are normalized and categorical features are encoded using one-hot
\vspace{-2mm}
\subsubsection{Baselines}
We select the following methods as baselines, 
which are commonly adopted by industry: 
Moving Average (MA), Extreme Gradient Boosting (XGB)~\cite{chen2016xgboost} which was deployed online previously, Random Forest (RF),  Linear regression (LR) and Auto Regressive Integrated Moving Average (ARIMA) are conventional time series methods. 
Deep neural network (DNN), Long short-term memory (LSTM), Sequence to Sequence (Seq2seq) and  Temporal Convolutional Network (TCN) are deep learning-based methods.
\vspace{-2mm}
\subsubsection{Metrics}
Two commonly used regression performance metrics are adopted to measure the performance of different models, including Mean Absolute Percentage Error (MAE) which measures the average absolute difference between the predicted results and the true values, and Mean Square Error (MSE) which is more sensitive to outlier points.
\vspace{-2mm}
\subsubsection{Model setting}
The batch size is set to 16, $T$ is set to 7. 
In the courier spatial learning component, layers of GCN is set to 1, and each GCN layer’s hidden and output dimension is 128. 
In the courier temporal learning component, layers of LSTM is set to 2, and each LSTM layer’s output dimension is 128. 
We also apply a dropout layer with a dropout rate of 0.1 to the LSTM. 
Our model is optimized using Adam and trained with 100 epochs as the performance can converge early 
with the learning rate set to 0.0001.

\subsection{Performance Comparison} 
We evaluate each approach for five independent rounds, and take the average of five rounds for each metric. 

As summarized in Table~\ref{performance}, we can observe that the DNN model performs the worst, while our DeepSTA shows the best performance. Basically, the performance of deep learning-based approaches is generally superior to that of conventional methods, 
demonstrating the superiority of deep learning-based methods in capturing temporal dependencies.
Additionally, among conventional methods, linear regression performs the worst, suggesting the existence of complex nonlinear relationships in features. 
Furthermore, among deep learning-based methods, DNN performs worst due to the lack of modeling non-linear relationships and spatial-temporal dependencies. 
Moreover, the performance of both LSTM and Seq2Seq is comparable but both are inferior to TCN.
This is because TCN is better at modeling long-term temporal dependencies in time series via dilated convolution module.
Overall, compared to all the methods above, our model achieves the best performance in terms of MAE and MSE, outperforming the best competitors
by 12.11\% in MAE and 13.71\% in MSE. 
Remarkably, none of the deep learning-based methods perform as well as DeepSTA. 
This is because the former mainly captures the temporal dependencies, while DeepSTA models the correlation between couriers and between road districts additionally and models the anomaly events separately.
\vspace{-3mm}
\begin{table}[htbp]
\centering
\caption{Performance of our model and baselines.}\label{performance}
\vspace{-2mm}
\begin{tabular}{|c|c|c|c|}
\hline
  \textbf{Method} & \textbf{Model} &  \textbf{MAE} & \textbf{MSE} \\
\hline
 & {MA} &  0.2272 & 0.0761 \\
 & {XGB}  &  0.1933 & 0.0584 \\
Conventional & {RF} &  0.1931 & 0.0579 \\
 & {LR} &  10.8885 & 2.8803 \\
 & {ARIMA} & 0.2174 &  0.0868 \\
 \hline
 & {DNN} & 26.0414  & 4.0381 \\
 Deep & {LSTM} &  0.1926 & 0.0576 \\
 learning & {Seq2seq} &  0.1909 & 0.0559\\
based & {TCN} &  \underline{0.1846} & \underline{0.0545} \\
 & {DeepSTA} &  $\mathbf {0.1593}$ & $\mathbf {0.0479}$ \\
 \hline
\multicolumn{2}{|c|}{{Improvements}} &  13.71\% & 12.11\% \\
\hline
\end{tabular}
\end{table}
\vspace{-6mm}
\subsection{Ablation Study}
To verify the effectiveness of each designed module, we produce several variants of the model. 
Specifically, we remove the features generated by Road district embedding  
denoted as "w/o Road district embedding". 
In the same way, 
"w/o GCN" removes the GCN  in Courier spatial learning,
"w/o LSTM" removes the LSTM in Courier temporal learning,
"w/o Memory" removes the memory network in Anomaly Pattern Attention,
"w/o RNN" removes the RNN in Anomaly learning and the external factors are concatenated with other features as model input,
"w/o Memory+RNN" removes both the memory network and the RNN.
\vspace{-3mm}
\begin{table}[htbp]
\centering
\caption{Ablation study results.}\label{Ablation}
\vspace{-4mm}
\begin{tabular}{|c|c|c|}
\hline
\textbf{Model} &  \textbf{MAE} & \textbf{MSE} \\
\hline
{w/o Road district embedding} &  0.1675 & 0.0503 \\
{w/o GCN} &  0.1860 & 0.0583\\
{w/o LSTM} &  0.1840 & 0.0552 \\
{w/o Memory} &  0.1766 & 0.0505\\
{w/o RNN} &  0.1785 & 0.0549 \\
{w/o Memory+RNN} &  0.1819 & 0.0568\\
{DeepSTA} &  $\mathbf {0.1593}$ & $\mathbf {0.0479}$\\
\hline
\end{tabular}
\end{table}
\vspace{-4mm}

As shown in Table~\ref{Ablation}, 
the DeepSTA outperforms its best competitors by 4.90\% in MAE and 4.77\% in MSE. It is clear that the performance of the model deteriorates when any of the components are missing. 
This confirms the validity of the design of each module and component in DeepSTA, demonstrating that the four types of information, \emph{i.e.}, the correlation between road districts, the correlation between couriers, the sequential dependencies and anomaly information, can contribute to the prediction performance.
\subsection{Parameter Setting}

\begin{figure}[htbp]
\includegraphics[width=0.9\linewidth]{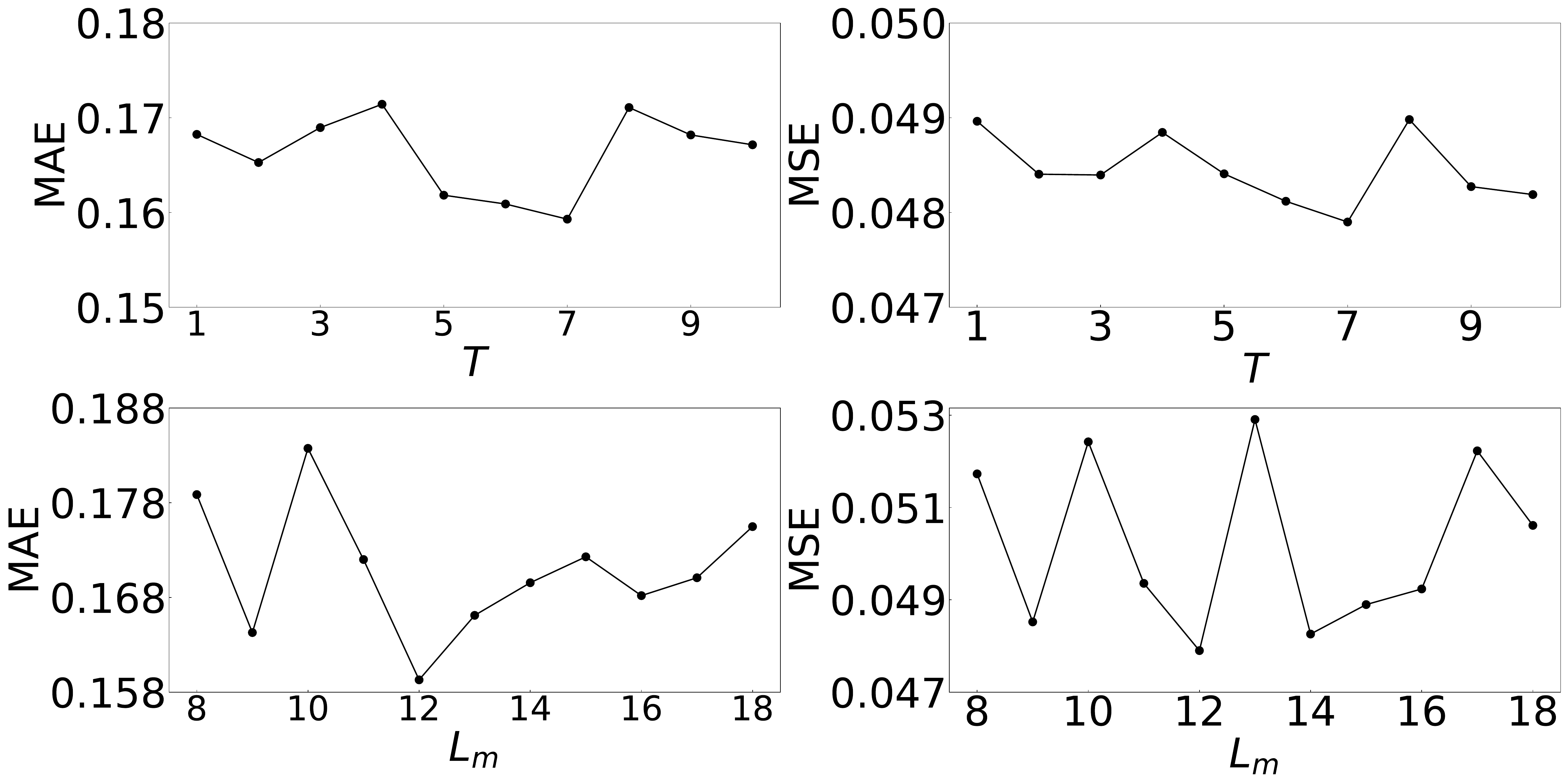}
\vspace{-4mm}
\caption{The impact of hyper-parameters.} \label{Tprevious}
\vspace{-6mm}
\end{figure}

In this section, we explore the impact of hyper-parameter of $T$, the length of historical timestamp and $L_m$, the number of patterns stored in the memory network. During the experiment, while changing one parameter, other parameters will be kept constant. The results are presented in Figure~\ref{Tprevious}.

First, we test the impact of $T$ which controls the amount of historical information by varying it from 1 to 10 at an interval of 1. 
The model achieves the best performance in all metrics when $T = 7$, indicating that the model's  performance will be impacted when input information is insufficient or redundant.
Besides, we modify $L_m$ ranging from 8 to 18. We can observe that the optimal performance is achieved when $L_m = 12$, suggesting that lacking or excessive external memory will lower model performance.
\vspace{-2mm}

\subsection{Case Study}
\begin{figure}[htbp]
\includegraphics[width=0.8\linewidth]{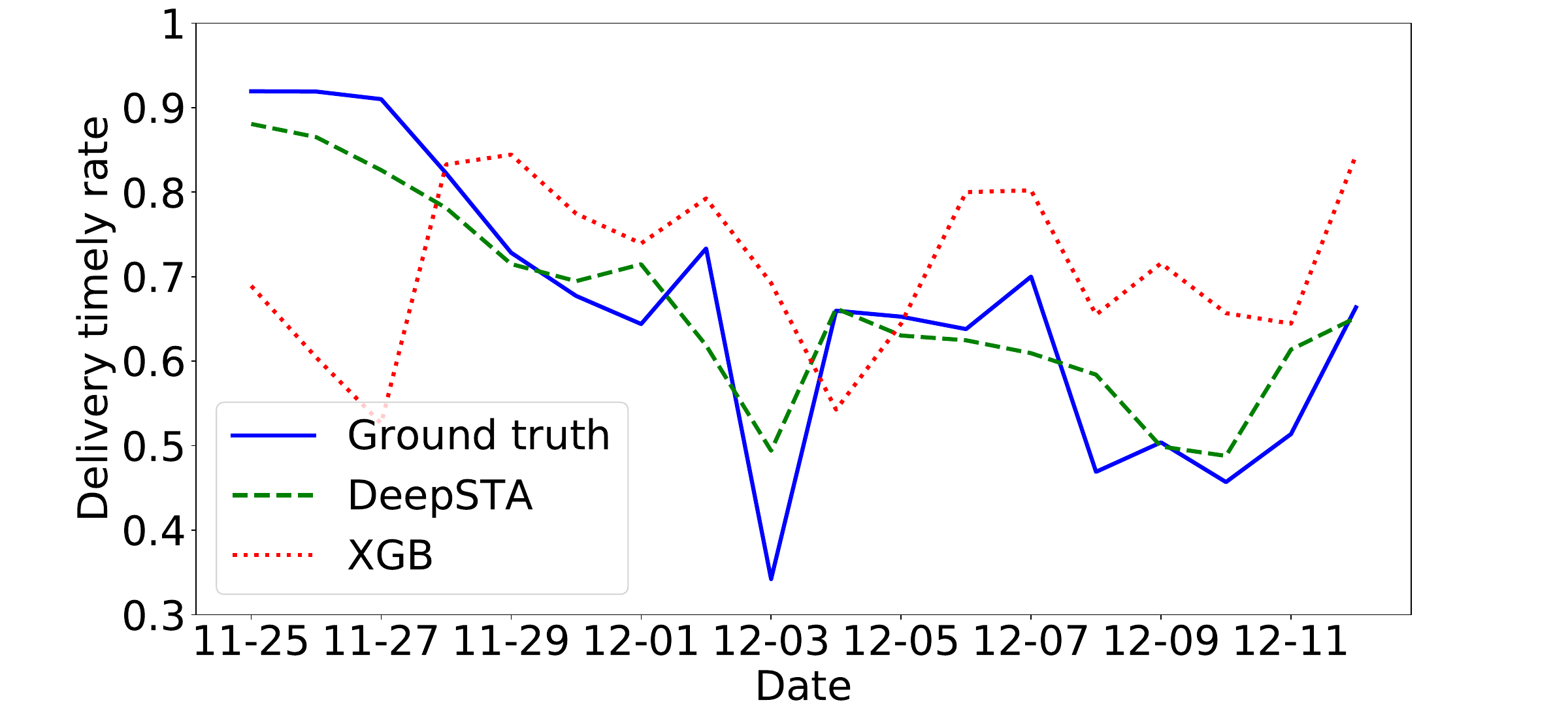}
\vspace{-4mm}
\caption{A case study of deployed models.} \label{case}
\vspace{-6mm}
\end{figure}

As shown in Figure~\ref{case}, we conduct a case study of a courier's timely rate prediction results using different methods. 
The figure reveals a time lag between the predicted values and actual values in the conventional Extreme Gradient Boosting (XGB) method, due to the model's reliance on the feature of the previous day's timely rate, 
while the DeepSTA performs better after the epidemic outbreak in late November, \emph{i.e.}, the DeepSTA outperforms XGB by 66.94\% in MAE and 86.99\% in MSE.
Incorporating our prediction results, the company assigned additional workers to assist couriers with low predicted results, which helped to minimize the COVID-19 outbreak impact on the express delivery in December 2022. 

\section{Conclusion}
In this paper, we propose a deep spatial-temporal attention model for couriers’ delivery timely rate prediction in anomaly conditions. 
The model captures the spatio-temporal dependencies of couriers as well as the impact of anomaly events. Further, it utilizes a memory network to tackle the lack of training data. 
The experiments on the real logistics dataset demonstrate the effectiveness of the model in the prediction task, and it has been deployed internally in JD Logistics to mitigate the risks of anomalous events.

\section{Acknowledgement}
This work was supported in part by Science and Technology Innovation 2030 - Major Project 2021ZD0114200, the National Nature Science Foundation of China under U22B2057, 62272260, 62272262, and China Postdoctoral Science Foundation under grant 2022M721833.

\bibliographystyle{ACM-Reference-Format}
\balance
\bibliography{main}










\end{document}